\documentclass[letterpaper, 10 pt, conference]{ieeeconf}

\IEEEoverridecommandlockouts                              % This command is only needed if 
\usepackage{hyperref}
\usepackage{blindtext}
\usepackage{graphicx}
\usepackage{multirow}
\usepackage{xcolor}
\usepackage{subcaption}
\usepackage[most]{tcolorbox}

\title{\LARGE \bf
BetterCheck: Towards Safeguarding VLMs\\for Automotive Perception Systems
}

\author{Malsha Ashani Mahawatta Dona, Beatriz Cabrero-Daniel, Yinan Yu, Christian Berger\\
\textit{University of Gothenburg} and \textit{Chalmers University of Technology}\\
Gothenburg, Sweden \\
\{malsha.mahawatta,beatriz.cabrero-daniel,christian.berger\}@gu.se, yinan@chalmers.se}

\begin{document}

\maketitle
\thispagestyle{empty}
\pagestyle{empty}
\newtcolorbox{boxA}{
    fontupper = \bf,
    boxrule = 1.5pt,
    colframe = black % frame color
}

%%%%%%%%%%%%%%%%%%%%%%%%%%%%%%%%%%%%%%%%%%%%%%%%%%%%%%%%%%%%%%%%%%%%%%%%%%%%%%%%
\begin{abstract}
Large language models (LLMs) are growingly extended to process multimodal data such as text and video simultaneously. Their remarkable performance in understanding what is shown in images is surpassing specialized neural networks (NNs) such as Yolo that is supporting only a well-formed but very limited vocabulary, ie., objects that they are able to detect. When being non-restricted, LLMs and in particular state-of-the-art vision language models (VLMs) show impressive performance to describe even complex traffic situations. This is making them potentially suitable components for automotive perception systems to support the understanding of complex traffic situations or edge case situation.
However, LLMs and VLMs are prone to hallucination, which mean to either potentially not seeing traffic agents such as vulnerable road users who are present in a situation, or to seeing traffic agents who are not there in reality. While the latter is unwanted making an ADAS or autonomous driving systems (ADS) to unnecessarily slow down, the former could lead to disastrous decisions from an ADS.
In our work, we are systematically assessing the performance of 3 state-of-the-art VLMs on a diverse subset of traffic situations sampled from the Waymo Open Dataset to support safety guardrails for capturing such hallucinations in VLM-supported perception systems. We observe that both, proprietary and open VLMs exhibit remarkable image understanding capabilities even paying thorough attention to fine details sometimes difficult to spot for us humans. However, they are also still prone to making up elements in their descriptions to date requiring hallucination detection strategies such as BetterCheck that we propose in our work.
\end{abstract}

\section{Introduction}

Nowadays, the adoption of Large Language Models (LLMs) can be seen in various domains including education, research, manufacturing, or healthcare~\cite{generativeAI_progress, healthcare11060887}. Applications of LLMs such as Pre-trained Transformers (GPT) in different domains, not limited to above mentioned fields, have enabled a positive influence opening new opportunities through their exceptional understanding and generation capabilities~\cite{generativeAI_progress}.

As modern vehicles grow into intelligent cyber-physical systems (CPS)~\cite{dona2024tapping} that are capable of housing powerful centralized processing units and hardware accelerators such as GPUs, executing specialized Neural Networks (NNs) has become increasingly feasible, which support Advanced Driver Assistant Systems (ADAS) and Autonomous Driving (AD). These advanced capabilities enable features such as real-time perception, decision-making, control functions, and even running LLMs locally without relying on cloud-based infrastructure backends~\cite{bmwCarExpert,bmwWebsite}.

\subsection{Problem Domain and Motivation} 

LLMs that work with multimodal data offer computer vision and natural language processing capabilities and are recently referred to as Vision Language Models (VLMs) when adopted for understanding image streams ~\cite{bordes2024introductionvisionlanguagemodeling}. These VLMs are designed to understand and generate text-based responses on visual inputs, including images and videos. Such state-of-the-art VLMs have shown remarkable capabilities in tasks such as image captioning, visual question-answering, and multimodal reasoning, demonstrating their potential in supporting perception and monitoring tasks in the automotive context~\cite{dona2024llms}. Due to exceptional natural language-based communication capabilities, they can also be used as Human Machine Interfaces (HMI) to support in-car passengers~\cite{hybridReasonongLLMinCars} and hence, make the vehicle much more accessible and inclusive. 

However, due to the potential risk of hallucinations~\cite{huang2023survey_hallucinations} generated by LLMs, the trustworthiness of such LLM-assisted systems and applications remains questionable. Therefore, when LLM assistance is used within safety-critical systems such as vehicles, it is important to design data processing pipelines embodying VLMs with safety guardrails to spot potential hallucinations and to mitigate them. 

\subsection{Research Goal and Research Questions}
\label{sec:researchQuestions}

The existing literature proposes hallucination detection strategies for LLM-assisted tasks. Manakul et al.~\cite{manakul2023selfcheckgpt} have evaluated and extended a technique called SelfCheckGPT that is capable of identifying nonsensical information generated by LLMs in text-based outputs. Dona et al.~\cite{dona2024llms} propose a variant of SelfCheckGPT that detects hallucination in multimodal contexts, specifically targeting automotive applications. The goal of our study is to determine the performance of the adapted SelfCheckGPT approach across different state-of-the-art LLMs when used for captioning images and for checking for hallucinations. In particular, we are focusing on the extent to which LLMs overlook traffic agents that may critically affect the trustworthiness of LLM-assisted perception and monitoring systems. 

% \begin{description}[leftmargin=!,labelwidth=\widthof{\bfseries RQ-1:}]
% \textbf{RQ-1:}Which LLM generates more correct captions? \textit{This is checked by analysing the human annotated ground truth data for each sentence in the generated captions.}

%\textbf{RQ-2:} Which model hallucinates more objects that are not in the image?

%\textbf{RQ-3:} Which model overlooks more agents?

%\textbf{RQ-4:} To what extent are the models providing relevant captions?

%\textbf{RQ-5:} To what extent can the models self-check their own captions?

% priposal of rqs
\textbf{RQ-1:} What is the quality, according to human evaluators, of state of the art VLMs in captioning real world automotive video footage?

\textbf{RQ-2:} To which extent do each of the tested VLMs hallucinate traffic agents or overlook them?

\textbf{RQ-3:} To what extent are the selected VLMs able to check their own results, and thus to discard captions likely to contain hallucinations or that overlook relevant traffic agents?

\subsection{Contributions and Scope}

We systematically assessed the adapted SelfCheckGPT hallucination detection framework across different combinations of LLMs with the goal to safeguard VLMs for automotive perception tasks. Our main contribution is the extension of the adapted SelfCheckGPT pipeline towards reducing chances of overlooking potential traffic agents and the systematic evaluation of VLMs as captioners and checkers. %, and the introduction of the union of sentences to the experiment pipeline, rather than restricting only to the sentences decomposed out of a single response $R_1$. 

\subsection{Structure of the Paper}

We structure the paper as follows: Section \ref{sec:relatedWork} reviews existing hallucination detection and mitigation strategies. Section \ref{sec:methodology} outlines the experiment pipeline for our study. The results and our interpretations are discussed in Section \ref{sec:results} and Section \ref{sec:AnalysisAndDiscussion}, respectively. We conclude the paper in Section~\ref{sec:conclusion}.

\section{Related Work}
\label{sec:relatedWork}

The adoptions and usage scenarios of SelfCheckGPT~\cite{manakul2023selfcheckgpt} were explored to identify gaps and limitations in the existing literature. In addition to that, selected studies were reviewed to understand their applicability in the automotive domain, specifically targeting LLM-assisted perception systems. 

Quite recently, Sawczyn et al.~\cite{sawczyn2025factselfcheckfactlevelblackboxhallucination} presented the hallucination detection technique FactSelfCheck that uses SelfCheckGPT as the basis. This method performs the hallucination detection at the fact level rather than on sentence or passage level. In this proposed technique, the facts are represented as a knowledge graph and later on, they are analyzed to check factual consistency. The sentence and passage level consistencies are calculated by aggregating the fact-level scores. According to the authors, the proposed technique outperforms SelfCheckGPT, which was used as its foundation. However, for this technique to be applicable in the automotive context, it should be adopted to deal primarily with image and video data, rather than only with text-based inputs. In addition to that, adding an intermediate layer that maps visual content into knowledge triplets (subject-predicate-object) could add extra weight and a barrier to the entire process, challenging the requirement of providing real-time hallucination detections. 

SelfCheckAgent~\cite{muhammed2025selfcheckagentzeroresourcehallucinationdetection} is another hallucination detection technique that refers to SelfCheckGPT as a baseline. This method combines multiple different agents to provide a multidimensional approach to detect potential hallucinations generated by LLMs. The authors have introduced three agents: a symbolic agent that assesses the factuality of the response, a specialized detection agent to spot hallucinations by using a fine-tuned transformer-based LLM, and a contextual consistency agent that exploits zero-shot and chain of thought prompting. While the contextual consistency agent can be adopted into the field of automotive upon addressing the limitations related to real-time applicability when chain-of-thought prompting is being used, the first two types of agents show limitations that hinder their applicability within the automotive domain. Overall, this method is limited to text-based data and does not incorporate multimodal data. % Therefore, the applicability of this framework to real-world scenarios involving safety-critical systems seems unclear at the moment.  

Yang et al.~\cite{yang2025hallucinationdetectionlargelanguage} propose a novel hallucination detection technique that exploits the metamorphic relations identified in the input text passages. The authors claim that the proposed method outperforms the SelfCheckGPT technique upon being evaluated under the same conditions. The proposed technique involves prompting an LLM to generate subsequent responses using synonyms and antonyms to the original response. A SelfCheckGPT-based consistency check will be carried out later to verify the factual consistency of each response. Inaccurate responses generated with synonyms and antonyms could lead to unreliable hallucination detections, even though the technique showcased better results for certain temperature settings of the models. Since this technique is based on antonyms and synonyms, there is a potential risk of introducing double negation and other semantically related issues that could potentially affect the overall performance. 

Dona et al.~\cite{dona2024llms} proposed a different adaptation of SelfCheckGPT for the automotive context. This technique explores the applicability of SelfCheckGPT for perception related tasks when prompted on image sequences. The authors recorded captions for image sequences, repeating the process multiple times, and checked the sentences of the first caption against the rest of the captions to assess whether each sentence is supported by the sequence of captions. Based on a sentence-level consistency score, the authors have implemented an exclusion mechanism to remove less consistent sentences from a response, ie., the sentences with low consistency are considered as potential hallucinations. In addition to that, the impact of different state-of-the-art large language models (LLMs), datasets, and lighting conditions have been explored, which strengthens the applicability of SelfCheckGPT within the automotive context. It has been observed that the choice of datasets made insignificant impacts on the final results, highlighting the fact that the proposed technique can be applied broadly, irrespective of the different driving behaviors. This technique has shown potential in being applied within the automotive context to detect and mitigate hallucinations when LLM-assisted perception and monitoring systems are used within vehicles.

\section{Methodology}
\label{sec:methodology}

\begin{figure*}
\centering
    \includegraphics[width=1\linewidth]{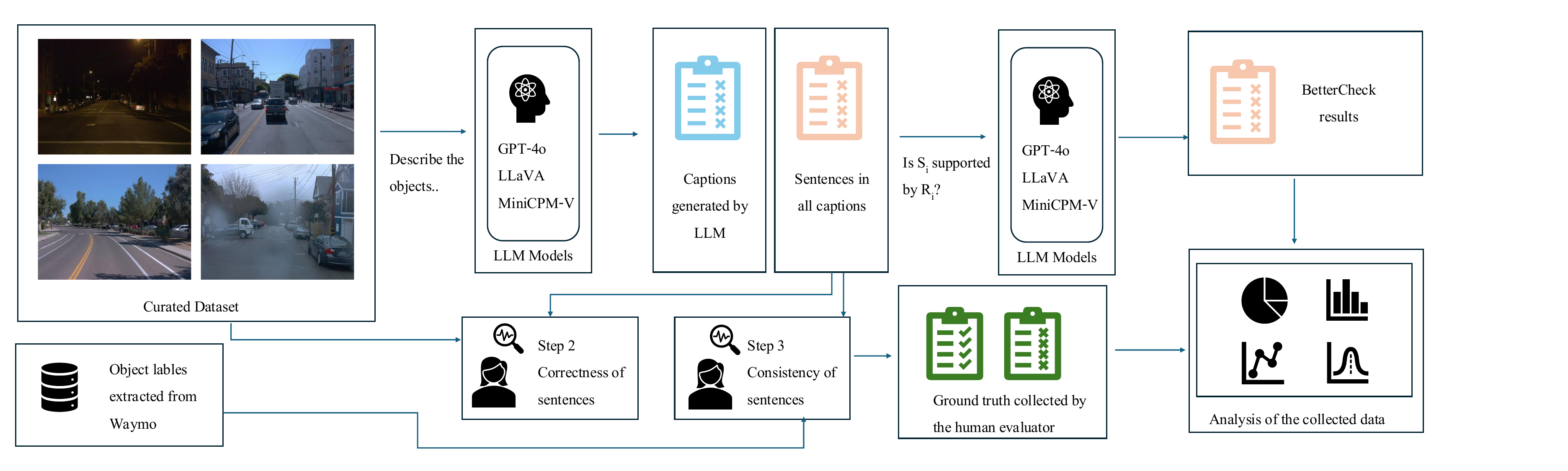}
    \caption{Overview diagram of the experimental setup: All images in the curated dataset are fed to the LLMs GPT-4o, LLaVA, and MiniCPM-V that we requested to describe the visible objects in each image using simple short sentences to allow for the LLMs'/VLMs' integration into a perception pipeline. The collected responses are evaluated by multiple LLMs against the ground truth human annotations to analyze the three experiments.}
    \label{fig:experimentPipeline}
\end{figure*}

The overall experiment pipeline is depicted in Figure~\ref{fig:experimentPipeline}. We selected 20 driving scenarios from a state-of-the-art dataset (Waymo) and extracted front camera images and the corresponding object labels. The selected images were fed into three state-of-the-art LLMs (GPT-4o, LLaVA, and MiniCPM-V) with a predefined prompt repeatedly to record the responses. The collected results were processed and statistically analyzed to answer the research questions.

\subsection{Dataset Curation and Preparation}

The Waymo Open Dataset~\cite{waymoDataset} collected in the USA is covering urban and suburban areas and was used when conducting this research study. It was collected in 2021 by Google, including 2030 segments that are approximately 20 seconds long. Each video frame contains data collected using five cameras, LiDAR, and radar sensors. The dataset covers various driving scenarios under different weather conditions, different day and night times, and neighborhoods, including urban and residential areas.  

\begin{figure*}
    \centering
    \includegraphics[width=1\linewidth]{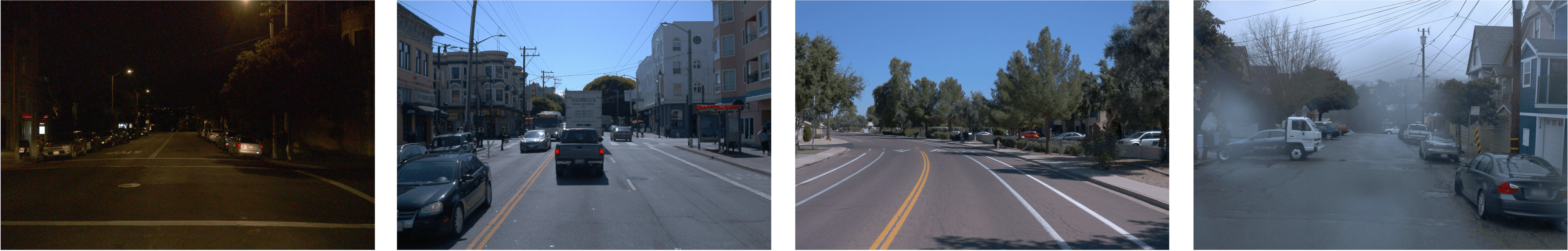}
    \caption{Example frames from the Waymo dataset.}
    \label{fig:1554232889558476_front}
\end{figure*}

For our experiment, we extracted images and their object labels by sampling every tenth frame, mainly focusing on front-facing camera images in the training set that are of 1920x1280 size in pixels. We curated a final dataset used for our study, including 500 images selected from 20 different driving scenarios depicting different neighborhoods, weather conditions, and day/night captures. From each selected driving scenario, 25 images were manually checked to ensure that each selected image differs from the previously selected images from the same scenario. This ensured to curate a diverse dataset that depicts different conditions, providing a broad spread in the sampled frames. Figure~\ref{fig:1554232889558476_front} indicates a few sample images selected from the Waymo dataset when curating our dataset.

\subsection{Data Collection}

As depicted in Figure~\ref{fig:experimentPipeline}, the curated dataset was fed into three state-of-the-art LLMs to collect their responses: We selected GPT-4o, a proprietary commercial model~\cite{openai2024gpt4ocard}, and two locally executed models, namely Large Language-and-Vision Assistant (LLaVA)~\cite{liu2023visualinstructiontuning}, and MiniCPM-V~\cite{yao2024minicpmvgpt4vlevelmllm}. We used the Ollama Python library~\cite{ollama2024} to access all three LLMs. The Ollama website \cite{ollama2024} contains all the necessary information regarding the token limits and the prompt lengths each model supports. 

As the first step in the experiment pipeline, we prompted each LLM to describe the different objects that are visible in each image. We designed the following prompt after several iterations of modifications to retrieve short sentences that explain only a single object at a time. We provided sample sentences within the prompt, following formal template-based prompting techniques to set clear expectations with the LLMs~\cite{schulhoff2025promptreportsystematicsurvey}. We followed the `Best of Three (BO3)' strategy~\cite{ronanki2022chatgpt} across all images by feeding the same image three times to each LLM, expecting more polished and coherent responses with less hallucinations. Each response $R_i$ was recorded and subjected to post-processing.  

\begin{boxA}
Describe the different objects visible in the image. Please write very simple and clear sentences. Use the format: ``There are [object].'' For example, ``There are cars. There are people. There are cyclists.'' Look carefully and make sure to mention all types of objects you see, especially people. If there are multiple types of objects in the image, provide a separate sentence for each type. 
\end{boxA}

As the next step of the study, each response $R_i$ was decomposed into sentence-level $s_{1 \dots n}$ elements. The individual sentences for all captions were evaluated one by one against the respective image by a group of human annotators to flag their respective correctness. Furthermore, each sentence $s_i$ was evaluated against the object labels extracted from the Waymo image labels by a group of human annotators to report the consistency of the sentences, ie., to check whether the LLM is missing any traffic objects that are visible in the image. This aspect of potential hallucinations is of particular highest importance when considering LLMs and VLMs for perception tasks as \emph{overlooking traffic agents that are present in a scene} can lead to disastrous consequences.

These correctness and consistency evaluation data were stored and later used as ground truth to evaluate the performance of the LLMs. Given that we involved multiple human annotators to label correct sentences and consistent sentences, it was crucial to use an inter-rater agreement to quantify how consistent the annotators were. Therefore, we created a 15\% overlap of captions randomly selected from each model's responses for the annotators and evaluated the human annotators' agreement using Cohen's kappa~\cite{cohen1960coefficient} to assess how much of the human-added annotations were consistent among the human annotators.

Next, we applied the adapted SelfCheckGPT~\cite{dona2024llms} to check how well another LLM can check the captions to detect hallucinations as a means to safeguard VLMs. In this step, all sentences $s_{1 \dots n}$ that belonged to all captions $R_{1 \dots n}$ were passed to the LLM one by one, along with each caption $R_i$ at a time, asking whether the sentence is supported by the respective caption. These binary \verb|Yes| or \verb|No| answers were recorded and analyzed statistically to understand how well the models were performing. 

\begin{boxA}
Context: {{CONTEXT}}  Sentence: {{SENTENCE}}
Is the sentence supported by the context above? Answer Yes or No:
\end{boxA}

In each instance, we used the same LLM, which was used as a captioner model before, as the checker model to assess the consistency in all instances, effectively allowing the VLMs to ``self check'' their own results~\cite{dona2024llms}. 

%So we will compare our assessments over 20 or 15% of the captions, that is ok. Those captions might represent 15-20% of the sentences or not. But it is a good chunk. Keep that in mind when it is time to write about that.

\subsection{Data Analysis}

The study data, including the generations by each of the VLMs as well as the human annotations, was retrieved and processed using the following steps:

\textbf{Step 1: Image captioning.} All the sampled images were prompted three times to each of the tested VLMs, together with the prompt described above in order to extract captions in a specific format. Said captions were stored in files, together with the execution time, to allow for later analysis. 

\textbf{Step 2: Correctness annotation.} In a second step, the human annotators were iteratively shown all sampled images captioned underneath with each of the sentences found within the captions. Sentences longer than 50 characters were discarded, as discussed above. The human was then asked to decide whether the sentence is either ``correct'' or ``incorrect'' with respect to the image.

\textbf{Step 2.1: Inter-rater agreement.} In this step, the results show that the inter-rater agreement ranges from 50\% to 80\% depending on the set of captions. This is partially due to some semantic differences between the words used by the VLMs. For instance, a ``pick up truck'' could be labeled as a vehicle, a car, or simply a ``truck,'' which led to differences in the judgment of the raters. We consider however that the agreement is substantial for the traffic agent categories present in the Waymo annotations.

\textbf{Step 3: Label consistency annotation.} The Waymo-provided labels were cross-checked by the human annotators in this step. To do so, the humans were shown the captions, sentence by sentence, together with all the labels corresponding to the image that the caption was generated for. The results were stored in order to study the confusion matrices and look for the percentage of overlooked traffic agents in subsequent steps.

\textbf{Step 4: Self-check.} All the sentences within the captions were programmatically paired with the two other captions for the image they were generated for in the prompt above. 

\textbf{Step 5: Statistical analysis.} Once all the generations and human annotations were gathered, a complex statistical analysis was performed. All the results of this analysis are reported in the following sections, including the percentage of correct sentences and captions for each model, statistical analysis for hallucinations (false positives) and overlooks (false negatives), a frequency word analysis of the generated sentences not concerned with the relevant traffic-agents, etc.

\section{Results} \label{sec:results}

\subsection{Description of Captions (Steps 1 and 2)} \label{sec:captiondescription}
%add sample captions for each model in boxes

The GPT4o-generated captions were accurate and precise based on their relevance with the prompt. The captions contained smaller sentences that explained exactly one object. In addition to that, the captions' length was moderate compared to the other two models. The model did not seem to overlook the traffic objects that were present, nor did it mention objects that were not present in the image. This nature significantly helped the analysis and human annotating tasks compared to the other models. 

\begin{boxA}
% Example of super straightforward GPT caption.
%GPT-4o caption for image 1554232889558476\_front.jpeg.
\textbf{GPT-4o : }``There are cars. There are buildings. There are streetlights. There are power lines. There are traffic signals. There are \textit{shops}. There are trees. There are sidewalks.''
\end{boxA}

MiniCPM-V showed exemplary capabilities of describing images. It could identify objects, wordings, an the like, even if they are appearing afar, making its image captioning capabilities exceptional. However, it did not closely follow the instructions provided in the prompt consistently. The model often generated long sentences that described a couple of objects, which made it difficult to annotate the sentences as correct or not, given that the sentences could be only partially incorrect. For instance, a single sentence could be explaining about a vehicle that is visible and a fire hydrant that is not visible in the image. Also it was noted that the responses generated by MiniCPM-V are poetic sometimes.  MiniCPM-V does not seem to miss anything, but it has many hallucinations.

\begin{boxA}
% Example of super complex minicimp caption for same image as the one for GPT.
%Part of a MiniCPM-V caption (for image 1554232889558476\_front.jpeg).
\textbf{MiniCPM-V : }``There's an SUV parked on a curb to our left. And another one in front of it, and then three more further down the road. It's \textit{all lined up like little soldiers}.''
\end{boxA}

The captions generated with LLaVA were often short and followed the prompt accurately, generating only short sentences that describe one object at a time. However, the model hallucinated a lot, especially with fire hydrants and parking meters that seemed to be everywhere. At the same time, LLaVA also managed to miss some objects when they were actually present in the picture.

\begin{boxA}
%Example of LLava caption mentioning a fire hydrant when it is not actually there.
%LLaVA caption for image 1554232889558476\_front.jpeg, mentioning parking meters, as usual.

\textbf{LLaVA : }``The sky is overcast. There are buildings along the street. There are cars on the road. There are \textit{parking meters} alongside the curb. There are several parked cars on the side of the street.''
\end{boxA}

\subsubsection{What are the captions that are not about the Waymo-annotated traffic agents?} \label{sec:step2}

The Waymo dataset provided object labels on \textit{{ Unknown, Vehicles, Pedestrians, Signs, and Cyclists }}. We compared the LLM-generated sentences against the Waymo-identified object labels to evaluate what critical traffic objects the models did overlook when describing the images. We analyzed the remaining sentences to understand what other objects the models were tempted to describe. 

\begin{figure*}
\centering
\begin{subfigure}[t]{0.3\textwidth}
\centering
    \includegraphics[width=.8\linewidth]{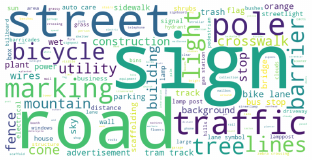}
    \caption{Words in GPT-4o captions that are correct but do not contain references to any of the Waymo-annotated traffic agents.}
    \label{fig:wordcloudgpt}
\end{subfigure}
\hfill
\begin{subfigure}[t]{0.3\textwidth}
\centering
    \includegraphics[width=.8\linewidth]{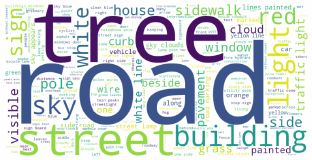}
    \caption{Words in MiniCPM-V captions that are correct but do not contain references to any of the Waymo-annotated traffic agents.}
    \label{fig:wordcloudminicpm}
\end{subfigure}
\hfill
\begin{subfigure}[t]{0.3\textwidth}
\centering
    \includegraphics[width=.8\linewidth]{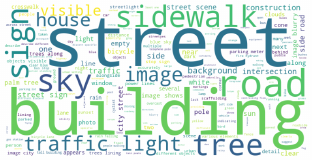}
    \caption{Words in LLaVA captions that are correct but do not contain references to any of the Waymo-annotated agents.}
    \label{fig:wordcloudllava}
\end{subfigure}
\caption{Wordclouds for the most common words in the correct sentences within the captions by each of the tested models, once removed the sentences mentioning the traffic agents annotated within Waymo.} \label{fig:wordclouds}
\end{figure*}

As seen in Figure~\ref{fig:wordcloudgpt}, GPT-4o often identified streets, street markings, lamps, hydrants, poles, structures, buildings, mailboxes, patches, posts, shadows, signposts, and streetlights. LLaVA often identified streets, buildings, signs, constructions, parking, buses, cones, corners, and poles as summarized in Figure~\ref{fig:wordcloudllava}. MiniCPM-V often identified roads, clouds, specific colors, poles, stop signs, wires, grass patches, lanes, and bicycles (though it seldom mentioned the cyclists, as prompted) as outlined in Figure~\ref{fig:wordcloudminicpm}.

\subsection{Quality of the captions (steps 2 and 3)} \label{sec:step3}

When evaluating the LLM-generated responses against the human-annotated ground truth for the correctness of each sentence, GPT4o's sentence-level correctness was 99.6\% whereas the caption-level correctness was 97.1\%.

MiniCPM-V's sentence-level correctness reached 94.8\% while its caption correctness remained 88\% . We excluded the sentences that exceeded the character limit of 50, considering MiniCPM-V's tendency to generate longer sentences. Only around 5\% of the short sentences generated by MiniCPM-V were incorrect. 

Llava's sentence correctness was 85.6\% and caption-level correctness was 71.9\% indicating the lowest results. 28,1\% of the captions generated by Llava contained one or more incorrect sentences.

\begin{figure}
    \centering
    \includegraphics[width=\linewidth]{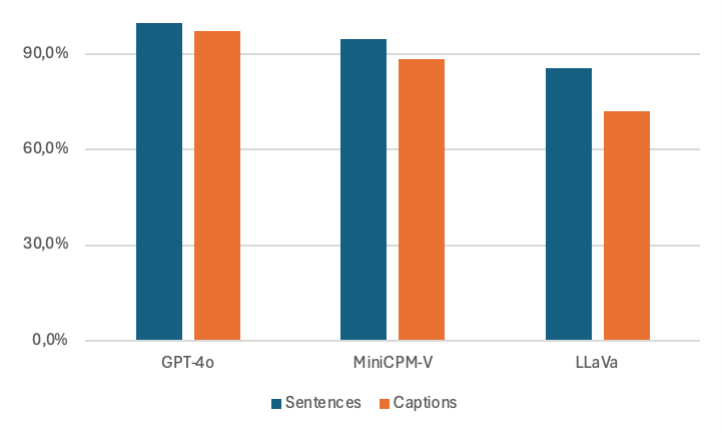}
    \caption{Bar chart showing the correctness of the sentences by each model (step 2).}
    \label{fig:sentenceandcaptioncorrectness}
\end{figure}

The Figure~\ref{fig:sentenceandcaptioncorrectness} indicates these results in a bar chart. 

When performance metrics were computed to statistically evaluate the models, GPT-4o achieved 100.0\% precision (0.0\% false positive rate), 78.04\% recall, 87.67\% F1 score, 88.2\% accuracy, 100.0\% specificity, and an overall MCC of 0.7885. 

MiniCPM-V achieved 100.0\% precision, 25.56\% recall (because of the 74.44\% false negative rate), 40.71\% F1 score, 60.0\% accuracy, 100.0\% specificity, and an overall MCC of 0.37.

LLaVA achieved 100.0\% precision, 56.41\% recall (43.59\% false negative rate), 72.13\% F1 score, 76.58\% accuracy, 100.0\% specificity, and an overall MCC of 0.61.

It makes sense that all models got 0 false positives, given that the Waymo annotations are manual and well-made. It is therefore impossible that the selected models find a pedestrian, a vehicle, or a cyclist that the humans did not spot. Thus, the 0.0\% false positive rate and 100.0\% precision. In this application case, however, as discussed by Dona et al.~\cite{dona2024llms}, the relevant metrics here are recall (to look for overlooking traffic agents) and MCC (to judge the overall performance of the models while compensating for class imbalance and the high precision).

\begin{figure*}
\centering
\begin{subfigure}[t]{0.3\textwidth}
    \includegraphics[width=\textwidth]{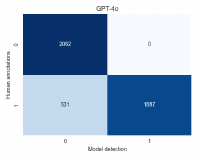}
    \caption{Confusion matrix showing the hallucinations and overlooks for GPT-4o sentences (against Waymo annotations).}
    \label{fig:confusiongpt}
\end{subfigure}
\hfill
\begin{subfigure}[t]{0.3\textwidth}
    \includegraphics[width=\textwidth]{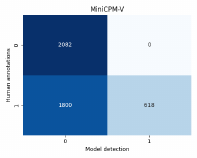}
    \caption{Confusion matrix showing the hallucinations and overlooks for MiniCPM-V sentences (against Waymo annotations).}
    \label{fig:confusionminicpm}
\end{subfigure}
\hfill
\begin{subfigure}[t]{0.3\textwidth}
    \includegraphics[width=\textwidth]{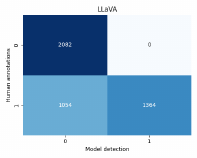}
    \caption{Confusion matrix showing the hallucinations and overlooks for LLaVA sentences (against Waymo annotations).}
    \label{fig:confusionllava}
\end{subfigure}
\caption{Confusion matrices for traffic agents in Waymo annotations (human) and detected by each of the models.} \label{fig:confusion}
\end{figure*}

\subsection{BetterCheck (step 4)} \label{sec:step4}

In~\cite{dona2024llms}, the authors evaluate the proposed adaptation of SelfCheckGPT for the automotive domain across different LLMs. We employed the same methodology under step 4 and checked the results to evaluate their performance. When sentence-level consistency was checked by prompting each sentence and the captions to an LLM, to check whether each sentence is supported by each response, we could compute the following metrics. Under this step, the same captioner model has been used to check their own results. For instance, GPT-4o-generated captions are checked by GPT-4o itself.  

GPT-4o achieved 99.72\% precision, 91.43\% recall, 95.4\% f1 score, 91.21\% accuracy, 40.91\% specificity, 59.09\% false positive rate, 8.57\% false negative rate. However, GPT-4o's MCC is only 0.07513.

MiniCPM-V achieved 100.0\% precision, 25.56\% recall (because of a 74.44\% false negative rate), 40.71\% F1 score, 60.0\% accuracy, 100.0\% specificity. The overall MCC was 0.3702.

LLaVA achieved 88.96\% precision, 88.81\% recall, 88.89\% F1 score, 80.64\% accuracy, 25.09\% specificity, 74.91\% false positive rate, 11.19\% false negative rate. However, LLaVA's overall MCC was only of 0.1384.

\begin{figure*}
\centering
\begin{subfigure}[t]{0.3\textwidth}
    \includegraphics[width=\textwidth]{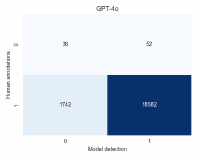}
    \caption{Confusion matrix for GPT-4o sentences (against human annotations).}
    \label{fig:selfcheckgpt}
\end{subfigure}
\hfill
\begin{subfigure}[t]{0.3\textwidth}
    \includegraphics[width=\textwidth]{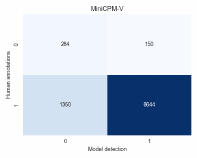}
    \caption{Confusion matrix for MiniCPM-V sentences (against human annotations).}
    \label{fig:selfcheckminicpm}
\end{subfigure}
\hfill
\begin{subfigure}[t]{0.3\textwidth}
    \includegraphics[width=\textwidth]{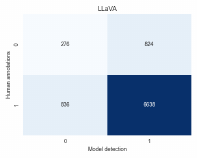}
    \caption{Confusion matrix for LLaVA sentences (against human annotations).}
    \label{fig:selfcheckllava}
\end{subfigure}
\caption{Confusion matrices for the human annotation on sentence correctness and the self-check by each of the models.} \label{fig:selfcheck}
\end{figure*}

\section{Analysis and Discussion} 
\label{sec:AnalysisAndDiscussion}

In order to address \textbf{RQ-1}, we compared the four models' ability to generate correct captions. This was done by analyzing the human-annotated ground truth data for each sentence in the generated captions. The results of this analysis show that GPT-4o is the model with the highest proportion of correct sentences within the captions, closely followed by MiniCPM-V, as seen in Figure~\ref{fig:sentenceandcaptioncorrectness}. Moreover, a comprehensive description of the captions by each of the models as well as their limitations, is provided in Section~\ref{sec:captiondescription}. 

In order to address \textbf{RQ-1}, we studied the word frequency in the captions that were annotated as correct by the human annotators but that did not mention any of the traffic agents annotated in Waymo. The goal was to understand to what extent are the models providing relevant captions. The results in Section~\ref{sec:step2}, and more specifically in Figure~\ref{fig:wordclouds}, show that VLMs are very capable of returning context-specific and relevant descriptions of the ego-vehicle's environment, though their generations might be difficult to programmatically interpret.

Also addressing \textbf{RQ-2}, we studied the amount of hallucinations, defined as detecting objects and traffic agents that are not in the image, that each model presented in the generated captions. The results in Section~\ref{sec:step3} show that all models are rather successful at generating image captions about relevant traffic agents and their environment (e.g., the weather and visibility conditions). At the same time, all the tested models also overlooked vulnerable road users in some occasions, as is summarised in Figure~\ref{fig:confusion}.

In a second step to address \textbf{RQ-2}, we also compared the generated captions to the Waymo-provided annotations for the images for relevant traffic agents. Section~\ref{sec:step3} compares the selected models in that regard, and discusses the trade-off that can be made between appropriate statistical metrics depending on the application case. For instance, recall might be the most relevant metric when it comes to identifying vulnerable road agents in front of the ego vehicle. MCC however, might highlight some vulnerabilities of the classifier, such as class imbalance.

Finally, in order to address \textbf{RQ-3}, we explore the correlation between the human annotation on sentence correctness and the self-check by each of the models, as described in Section~\ref{sec:methodology}, following the work by Manakul et al.~\cite{manakul2023selfcheckgpt} and Dona et al.~\cite{dona2024llms}. The results clearly show, in line with Dona et al.~\cite{dona2024llms}, that the selected models can check their own generations to improve the overall correctness of the image captions by removing sentences that are not consistent across prompt repetitions, as presented in Section~\ref{sec:step4}, and more specifically in Figure~\ref{fig:selfcheck}.

%
% threats to validity
\section{Conclusion and Future Work}
\label{sec:conclusion}

State-of-the art vision language models (VLMs) show remarkable performance when processing multimodal data in various application domains. In our work, we have systematically assessed the performance of 3 VLMs, GPT-4o, LLaVA, and MiniCPM-V on a curated subset of the Waymo Open Dataset to evaluate two crucial quality parameters: To what extent to VLMs overlook traffic agents that are present in a traffic scence, and to what extent are VLMs prone to describe traffic agents that are not there in reality. While the latter may lead to overly defensive driving of an autonomous driving system (ADS), facing the former may lead to disastrous decisions of an ADS in reality when not braking the vehicle in a critical traffic situation.

We observed that even both, the proprietary LLM GPT-4o and the open LLM MiniCPM-V show exceptional performance in describing a given traffic situation at hand. Yet, the latter model is still more prone to see more things that are actually not present or are located somewhere else in the analyzed image than reported in its response. Our results show that LLMs' and VLMs' performance allows their applicability for feature detection and extraction in automotive contexts complementing state-of-the-art specialized neural networks (NNs) like Yolo, which have a very limited vocabulary, ie., known objects that they can find in images, compared to LLMs.

However, we also conclude that safeguarding LLMs and VLMs that are part of a perception pipeline in automotive systems is unavoidable to lower the chances for facing issues with the two, aforementioned quality parameters. We have proposed BetterCheck as an adaption to the state-of-the-art hallucination detection technique SelfCheckGPT, where a combination of VLMs and LLMs are used to assess jointly a VLM's response to an image understanding prompt. While the results are pointing into the right direction, today's state-of-the-art VLMs are still too resource intense, either from a computational perspective when executed locally, or from the network round-trip-time when accessed in a cloud, to be considered for deployment in an ADAS or AD. However, we expect that next generations of these models will further address these aspects and hence, making them a potentially valuable addition to automotive perception systems. Furthermore, better prompting techniques to tailor an LLM's or VLM's output to be suitable for a perception system with multiple components are needed finding the right balance of allowing these models to be as descriptive and detailed as possible, while providing them certain guiding to make them compatible for automated processing that may require a restricted vocabulary.

\section*{Acknowledgments}
This work has been supported by the Swedish Foundation for Strategic Research (SSF), Grant Number FUS21-0004 SAICOM, Swedish Research Council (VR) under grant agreement 2023-03810, and the Wallenberg AI, Autonomous Systems and Software Program (WASP) funded by the Knut and Alice Wallenberg Foundation.

\bibliographystyle{ieeetr}
\bibliography{reference}

\end{document}